\documentclass[sigconf]{acmart}
\AtBeginDocument{%
  }

\setcopyright{acmlicensed}
\copyrightyear{2024}
\acmYear{2024}
\acmDOI{XXXXXXX.XXXXXXX}

\acmConference[SA '24]{SIGGRAPH Asia}{December 03--06,
  2024}{Tokyo, Japan}

\acmISBN{978-1-4503-XXXX-X/18/06}

\usepackage{hyperref}
\usepackage{listings}
\usepackage{courier} 
\begin{document}

\title{Can GPTs Evaluate Graphic Design Based on Design Principles?}

\author{Daichi Haraguchi}
\affiliation{%
  \institution{CyberAgent}
  \country{Japan}
}

\author{Naoto Inoue}
\affiliation{%
  \institution{CyberAgent}
  \country{Japan}
}

\author{Wataru Shimoda}
\affiliation{%
  \institution{CyberAgent}
  \country{Japan}
}

\author{Hayato Mitani}
\affiliation{%
  \institution{Kyushu University}
  \country{Japan}
}

\author{Seiichi Uchida}
\affiliation{%
  \institution{Kyushu University}
  \country{Japan}
}

\author{Kota Yamaguchi}
\affiliation{%
  \institution{CyberAgent}
  \country{Japan}
}

\renewcommand{\shortauthors}{Haraguchi et al.}

\begin{abstract}
Recent advancements in foundation models show promising capability in graphic design generation.
Several studies have started employing Large Multimodal Models (LMMs) to evaluate graphic designs, assuming that LMMs can properly assess their quality, but it is unclear if the evaluation is reliable.
One way to evaluate the quality of graphic design is to assess whether the design adheres to fundamental graphic design principles, which are the designer's common practice.
In this paper, we compare the behavior of GPT-based evaluation and heuristic evaluation based on design principles using human annotations collected from 60 subjects.
Our experiments reveal that, while GPTs cannot distinguish small details, they have a reasonably good correlation with human annotation and exhibit a similar tendency to heuristic metrics based on design principles, suggesting that they are indeed capable of assessing the quality of graphic design.
Our dataset is available at \url{https://cyberagentailab.github.io/Graphic-design-evaluation/}.

\end{abstract}



\begin{teaserfigure}
  \includegraphics[width=\textwidth]{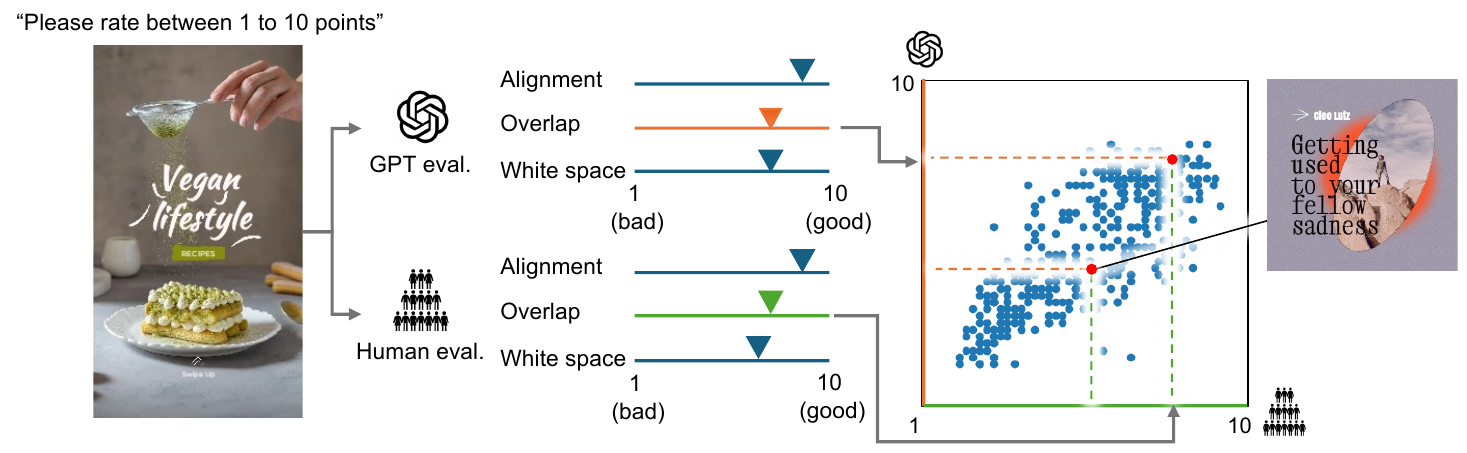}\\[-5mm]
  \caption{Overview of our study. We investigate GPT-based evaluation ability for graphic designs in three design principles ``alignment,'' ``overlap,'' and ``white space.''}
  \Description{Overview of our study.}
  \label{fig:teaser}
\end{teaserfigure}

\maketitle

\section{Introduction}
Foundation models learn from a large-scale corpus and exhibit remarkable generalization capability across various tasks, and the same is true for graphic design tasks~\cite{chen2023textdiffuser,graphist2023hlg,jia2023cole,inoue2024opencole}.
While successful in certain graphic design tasks, it is still not apparent whether foundation models, such as GPT-4o, can reliably judge the quality of graphic design.
Generally, high-quality graphic designs tend to follow design principles that are the designer's common practices, such as alignment or repetition, as described in~\cite{graham2002basics,robin2014non-designer}.
While humans can judge whether a given design follows these principles, human evaluation is time-consuming and not scalable.
One of the early attempts in automatic evaluation is \cite{o2014learning}, which introduces hand-crafted metrics for optimization.
Recent studies of graphic design generation have employed Large Multimodal Models (LMMs), particularly GPT-4~\cite{achiam2023gpt}, to directly estimate the quality.
However, these studies do not consider the viewpoints of design principles for the evaluation. 

In this paper, we quantitatively study the behavior of GPTs with respect to human evaluators in assessing graphic designs, as illustrated in Figure~\ref{fig:teaser}.
To investigate the performance of GPTs, we employ heuristic metrics as baseline.
We compare these approaches across three representative design principles, alignment, overlap, and white space, by manipulating graphic designs.
We curate a dataset of graphic banner designs from an online service and perturb original designs to artificially generate low-quality designs for evaluation.
Then, we ask human subjects to rate those designs in terms of the design principles.
This human rate annotation allows us to assess which method -- heuristic evaluation or GPTs -- better aligns with human evaluation.
We also discuss the qualitative comparison of heuristic and GPT-based evaluation.

We summarize our contributions in the following.
\begin{itemize}
    \item We empirically study the relationship between hand-crafted and GPT-based evaluation for graphic designs with respect to human evaluation. 
    \item We build a human-rated dataset of graphic designs that have varying degrees of quality to study the design quality.

    \item We find a higher correlation between human annotation and GPTs than heuristic metrics, suggesting that GPTs can give reliable judgment for graphic design quality under certain conditions.
\end{itemize}

\section{Graphic design principles}\label{sec:design_principles}
\begin{figure}[t]
  \centering
  \includegraphics[width=\linewidth]{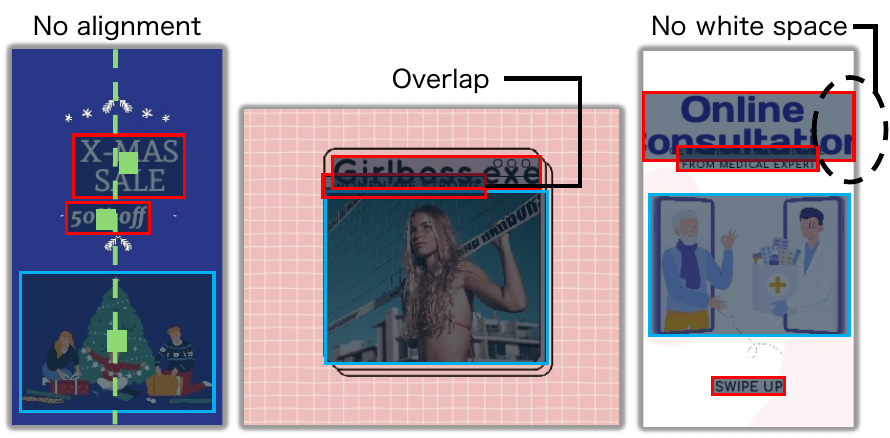}\\[-3mm]
  \caption{Negative examples of three design principles. }
  \Description{This is a figure.}
  \label{fig:principle}
\end{figure}
Graphic design principles are the designer's common practices to create aesthetic work~\cite{graham2002basics,robin2014non-designer}.
In this paper, we employ representative design principles used in several studies of graphic design generation~\cite{o2014learning,kong2022aesthetics++}: \emph{alignment}, \emph{overlap}, and \emph{white space}.
Alignment and overlap are commonly used in evaluation for layout generation, such as ~\cite{li2020attribute}. 
White space is also a critical factor in graphic design generation approaches~\cite{kong2022aesthetics++}.
We follow three design principles based on~\cite{o2014learning}.
Below, we briefly describe each principle.

\paragraph{Alignment}
We consider the arrangement of elements such that their edges line up along common rows or columns to express a sense of order and structure.
\begin{enumerate}
    \item Alignment along with the horizontal and vertical direction is considered.
    \item The elements that align at a glance but slight misalignment are penalized because it is visually displeasing.
    \item Larger alignment groups (i.e., aligned elements distant from each other) are preferred as they produce simpler designs with more unity between elements.
\end{enumerate}

\paragraph{Overlap}
Inappropriate overlap reduces readability. We consider the following aspects.
\begin{enumerate}
    \item The three types of overlap, the overlap of elements on text, the overlap of text on graphics, and the overlap of graphics on other graphics, are considered.
    \item Hard-to-read text because of insufficient color contrast between a text and the background color is penalized.
    \item The graphic design that includes elements extending past the boundaries is also penalized.
\end{enumerate}

\paragraph{White space}
White space is for the appropriate amount of space in a design for better readability.
\begin{enumerate}
    \item A large ratio of white space that is not covered by design elements (e.g., graphics and tests) is preferred.
    \item However, the graphic design with a too large region of empty white space on the image is undesirable.
    \item The greater distance between each element is preferred.
    \item Uniformed vertical spacing of each text element is preferred.
    \item Wider border margins (i.e., the white space at the edges of the image) for each element are preferred.
\end{enumerate}

For better understanding, we show some visually unappealing examples because they violate the principles in Figure~\ref{fig:principle}.

\section{Evaluation approach}
In this paper, we compare the evaluation metrics of the heuristic approach and the GPT-based approach.
In both approaches, we input a graphic design and then obtain the score of the input.
We set the lower and upper bounds for both metrics, where the higher values indicate better quality.

\subsection{Heuristic evaluation metrics}
We employ the hand-crafted formulation in~\cite{o2014learning} as heuristic evaluation metrics, which were originally designed for graphic design optimization.
We adopt the formulas related to the alignment, overlap, and white space as evaluation metrics.
These metrics directly consider the size and coordinates of text or other graphic elements.
We normalize the metric range from 0 to 1.

\subsection{GPT-based evaluation metrics}
\begin{figure}[t]
  \centering
  \includegraphics[width=\linewidth]{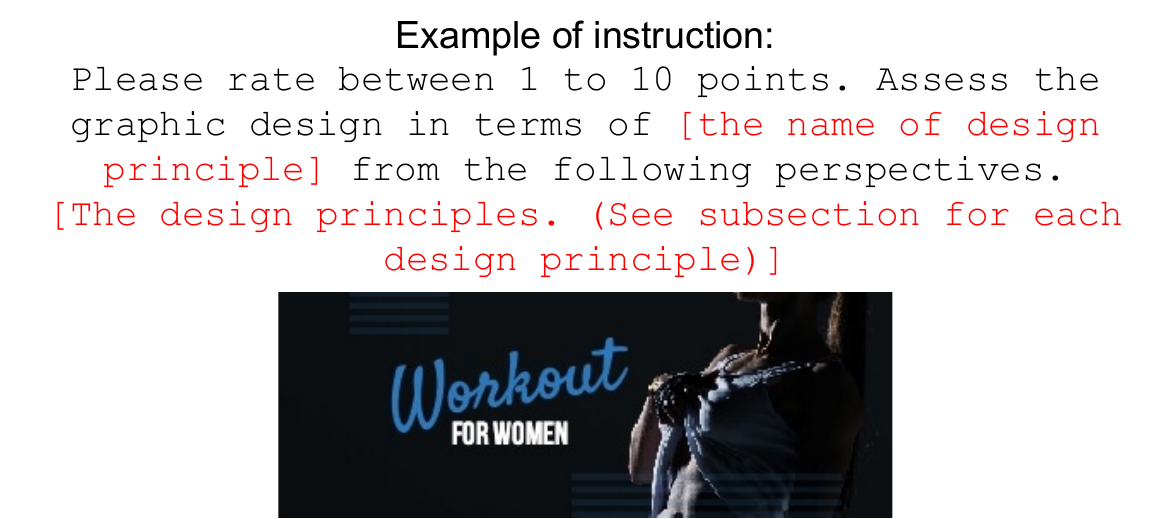}\\[-3mm]
  \caption{How to rate graphic designs by GPT and humans. The detailed input prompts for the GPT is described in the Appendix.}
  \label{fig:human_eval_eg}
  \Description{Annotation example.}
\end{figure}
We give prompts to GPT-4o~\footnote{\url{https://openai.com/index/hello-gpt-4o/}} and ask for scores in terms of alignment, overlap, and white space.
We made the prompt based on explanations of design principles and formulas described in~\cite{o2014learning}.
See the Appendix for more details.
We render and rasterize the original design to create an input prompt.
We show an example in Figure~\ref{fig:human_eval_eg}, where GPT-4o gives scores from 1 to 10 to the input based on a specific aspect.

\section{Dataset}
We collect graphic design templates from VistaCreate
~\footnote{\url{https://create.vista.com}}, which hosts a large number of banner or poster designs.
The templates contain coordinate and size information for graphic and text elements within a design.
We randomly sampled one hundred templates from VistaCreate and perturbed coordinate and size parameters to create aesthetically inferior samples.
We apply two kinds of perturbations in our experiments: $x$-coordinate and font size.
We perturb the $x$-coordinate of the text position to evaluate alignment and the font size of text elements to evaluate overlap and white space.
We give three ranges of perturbation, small, medium, and large, to investigate the sensitivity of metrics.
We combined the 100 original samples and 600 perturbed samples to build a dataset of 700 samples.

For each design, we collect human scores for alignment, overlap, and white space.
We recruited 60 participants via crowdsourcing and collected five annotations per graphic design, where we asked participants for scores in the 1 to 10 range, as shown in Figure~\ref{fig:human_eval_eg}.
We use the average score of five annotations in our experiments.

\section{Experiments}
\begin{figure}[t]
  \centering
  \includegraphics[width=\linewidth]{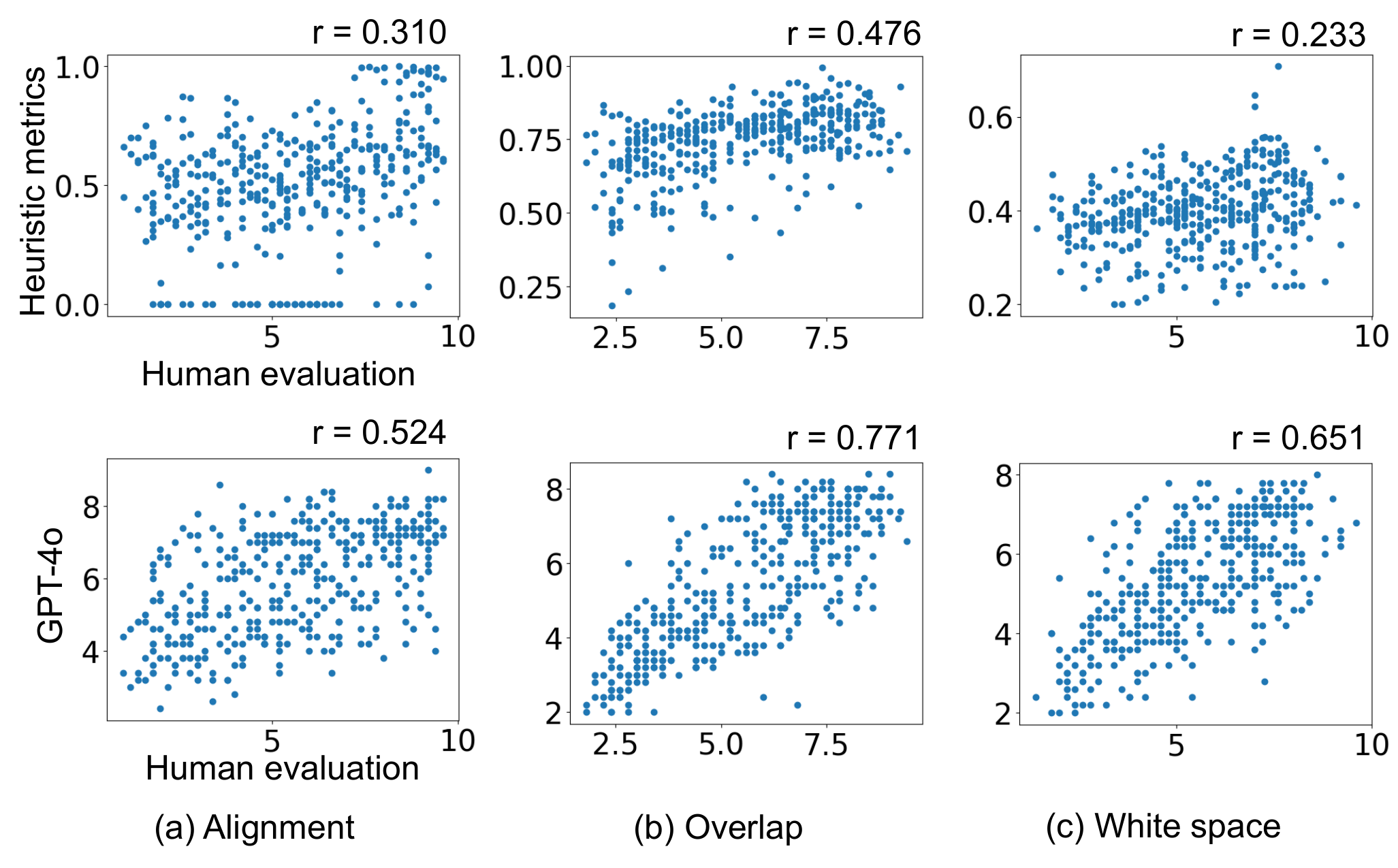}\\[-3mm]
  \caption{Correlation between human annotation and heuristic or GPT scores. The $r$ is the Pearson correlation coefficient.}
  \Description{This is a figure.}
  \label{fig:yy_plot}
\end{figure}
\subsection{Quantitative evaluation}\label{sec:quantitative}

\noindent\textbf{Setup}
For a fair comparison with human evaluation and detailed analyses, we conduct the GPT-based evaluation five times and use the average score for the GPT score.
We set a sampling temperature, which controls the randomness of the GPT, to 1 (default value).
Higher temperature (e.g., 0.9) shows more diverse output.
Therefore, our experimental setting takes the diversity into account.

\noindent\textbf{Correlation to human evaluation}
To analyze the correlation between human annotation and the scores by automatic evaluation, we prepare two types of scatter plots of evaluation scores: one comparing heuristic metrics with human evaluation and the other comparing GPT-4o with human evaluation for each design principle, as shown in Figure~\ref{fig:yy_plot}.
Interestingly, GPT-4o shows a better correlation with human evaluation than heuristic metrics.
This may be because heuristic metrics are designed to optimize a graphic design by comparing the graphic design before and after optimization but not for comparing two totally different designs.

\begin{figure}[t]
  \centering
  \includegraphics[width=\linewidth]{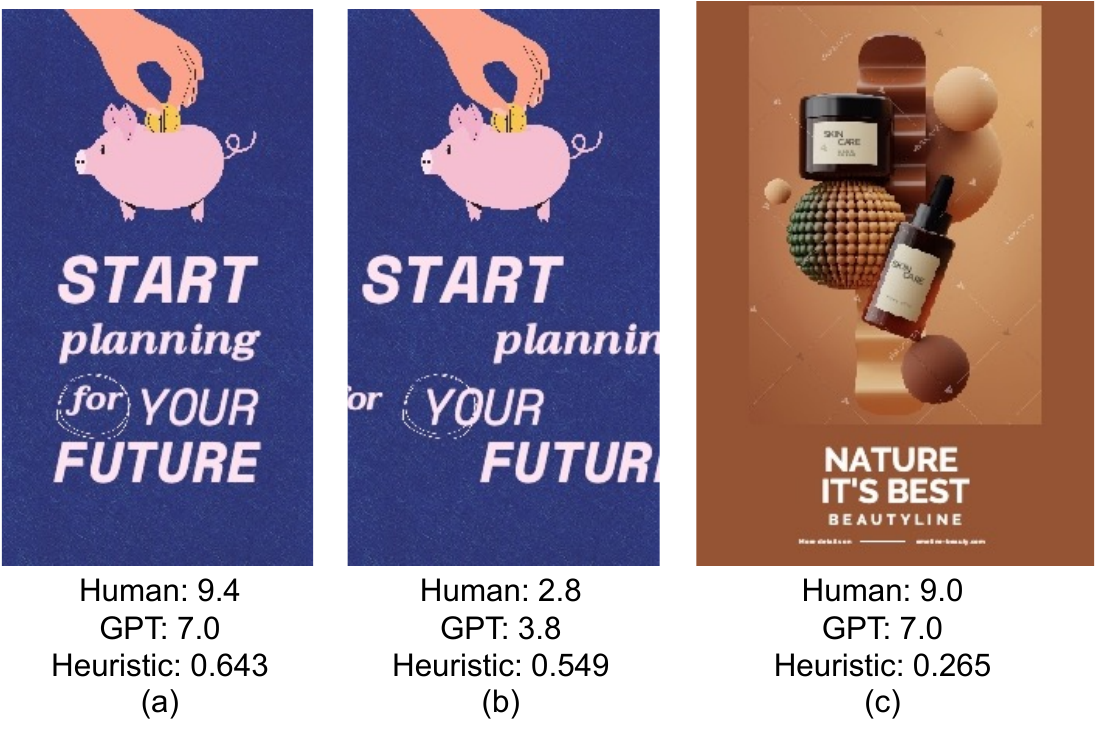}\\[-4mm]
  \caption{Graphic designs with their alignment scores.}
  \Description{This is a figure.}
  \label{fig:diff_score}
\end{figure}

We show an example of completely different graphic designs and their scores in Figure~\ref{fig:diff_score}.
According to the heuristic metrics, the order of better design is (a), (b), and (c). However, the order according to GPT-4o and human evaluation is (a), (c) (or (a) equal (c)), and (b). The heuristic evaluation quite degrades the score between (a) and (c) despite the fact that both graphic designs are created by human designers.
On the other hand, GPT-4o and human evaluation are not much different.
The results suggest that GPT-4o could be a suitable approach for quality evaluation for recent LMM-based generators since they generate significantly different designs by altering sampling parameters (or random seeds).

\begin{figure}[t]
  \centering
  \includegraphics[width=\linewidth]{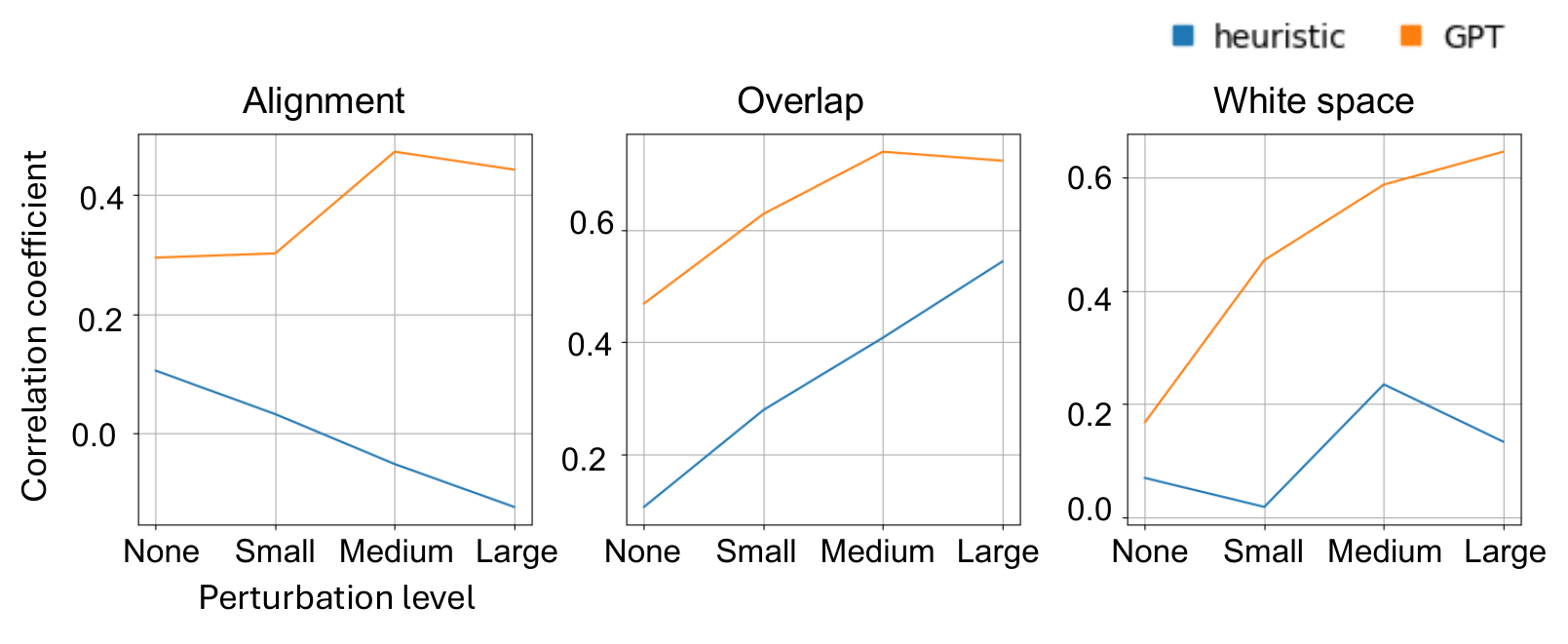}\\[-4mm]
  \caption{Correlation coefficient of the scores between human evaluation and each method.}
  \Description{This is a figure.}
  \label{fig:correlation}
\end{figure}

\begin{figure}[t]
  \centering
  \includegraphics[width=0.75\linewidth]{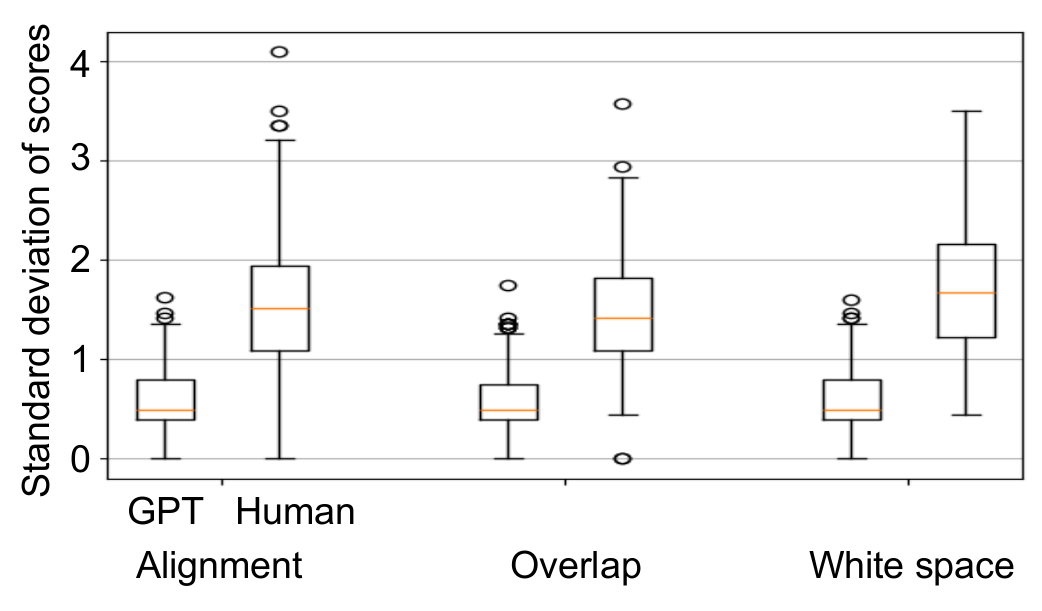}\\[-4mm]
  \caption{Box plot of the standard deviation of scores.}
  \Description{This is a figure.}
  \label{fig:boxplot}
\end{figure}

\noindent\textbf{Sensitivity}
We investigate whether GPT-based evaluation is effective across designs of diverse qualities by calculating the correlation coefficients for each perturbation level, as shown in Figure~\ref{fig:correlation}.
GPT scores exhibit a stronger correlation with human annotation compared to heuristic scores across all metrics and perturbation levels.
Additionally, the correlation of GPT scores tends to increase as perturbation levels rise.
This implies that when perturbations are significant, such as in the case of a graphic design with a noticeably poor appearance, achieving a stable evaluation becomes easier and more efficient than humans.

\noindent\textbf{Reliability}
We investigate how stable the GPT scores are since running GPT multiple times may give different scores.
We show the standard deviation of the GPT 
scores and human annotation in Figure~\ref{fig:boxplot}. 
The standard deviation of GPT scores is lower than that of the human evaluation scores.
This result suggests that GPT scores with a single run for each principle are practical and cost-effective evaluation metrics.

\subsection{Qualitative analysis}
\begin{figure}[t]
  \centering
  \includegraphics[width=\linewidth]{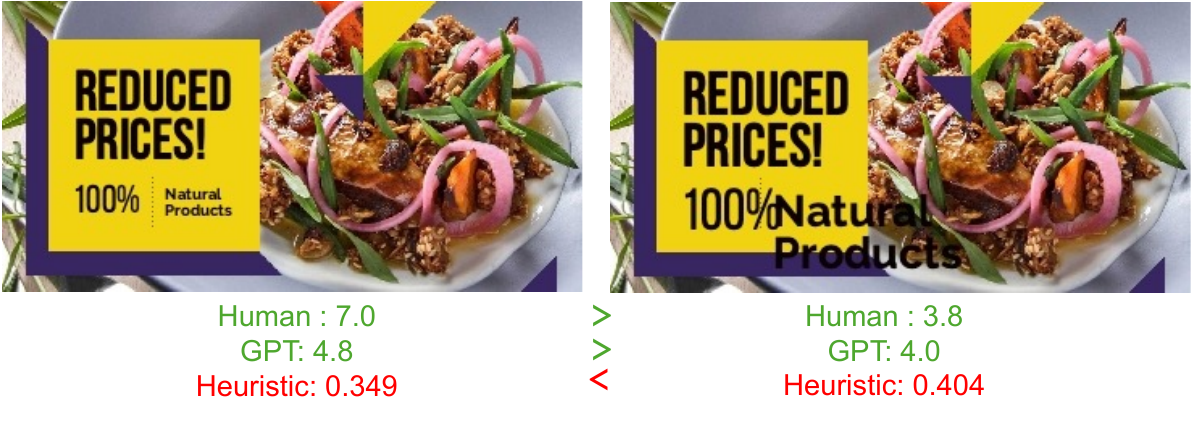}\\[-4mm]
  \caption{A sample with a correct white space assessment by GPT.}
  \Description{This is a figure.}
  \label{fig:incorrect_od}
\end{figure}

We show typical cases where GPT-based evaluation succeeds while heuristic evaluation fails and vice versa.
We show cases where only GPT-based evaluation succeeds in~Figure~\ref{fig:incorrect_od}.
The design includes objects in the background, but heuristic evaluation considers the background as white space.
Heuristic evaluation is difficult when assessing graphic designs that include an object embedded in the background.

\begin{figure}[t]
  \centering
  \includegraphics[width=\linewidth]{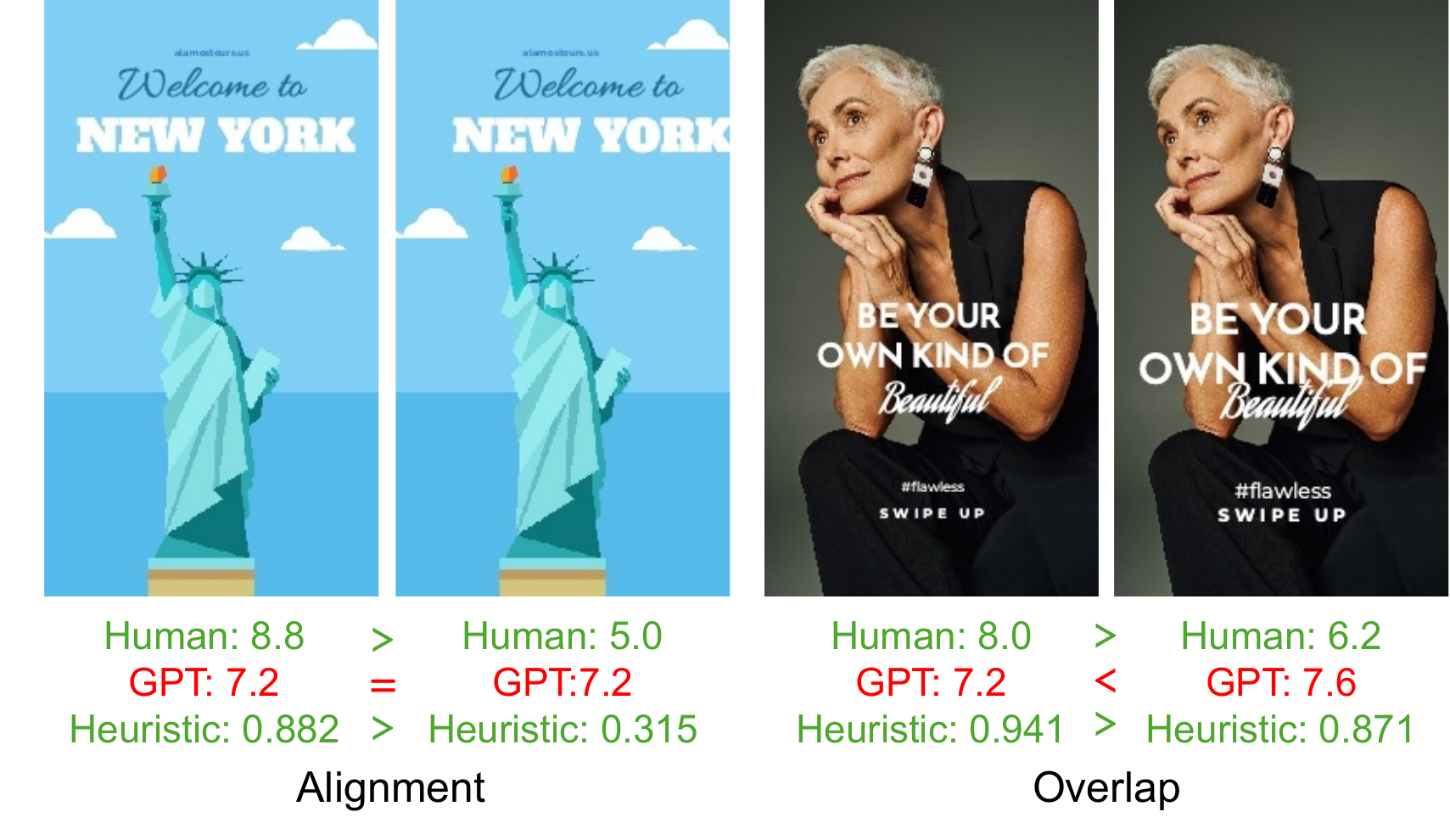}\\[-4mm]
  \caption{The examples of correctly assessed samples by heuristic evaluation.}
  \Description{This is a figure.}
  \label{fig:incorrect_gpt}
\end{figure}

We also show cases where only heuristic evaluation succeeds in~Figure~\ref{fig:incorrect_gpt}.
Since heuristic metrics are vector-based, they can capture slight differences in the design directly from the vector values.
However, GPT-based evaluation struggles to detect such differences accurately.
A similar limitation of GPT has also been reported in another study~\cite{you2023ferret}.

\section{Conclusion and Future Work}
In this paper, we investigated the appropriateness of using heuristic evaluation metrics and GPT-4o for evaluating graphic designs focusing on design principles. 
To achieve this, we collected the large-scale human annotation and analyzed the correlation between the annotation and each evaluation metric.
Our experiment showed that GPT-based evaluation better correlates with human annotation.

For future work, we will jointly evaluate several design principles to assess the overall goodness of graphic designs.
Additionally, we plan to evaluate graphic designs beyond design principles, such as font choices.

\begin{acks}
A part of this work was supported by JST ACT-X (Grant No. JPMJAX22AD).
\end{acks}
\bibliographystyle{ACM-Reference-Format}
\bibliography{ref}

\appendix

\section{Details of design principles}\label{sec:principle}
We summarize the design principles used in our experiments.
These design principles are based on \cite{o2014learning}.
\subsection{Alignment}
This metric assesses the alignment of graphic and text elements and consists of three major components. 
First, it assesses whether there is alignment between neighboring elements. Second, it evaluates the alignment error between elements intended to be precisely aligned within a neighborhood. 
If the elements appear to be aligned but exhibit slight misalignment, the evaluation is based on the degree of this misalignment. 
Third, the metric considers alignment among distant elements. If the elements between two separated elements are aligned with these two elements, both distant elements are also considered aligned and are evaluated accordingly.

We transcribe the alignment principle to the prompt below.

\begin{lstlisting}
Correct alignment is an important aspect of design that has been modeled in other layout applications. 
Text and graphic elements are aligned on the page to indicate organizational ructure and aesthetics. 
Please evaluate the alignment of the input graphic design considering the following points.
1. Alignment along with the horizontal and vertical direction is considered.
2. The elements that align at a glance but slight misalignment are penalized because it is visually displeasing.
3. Larger alignment groups (i.e., aligned elements that are distant from each other) are preferred as they produce simpler designs with more unity between elements.
\end{lstlisting}

\subsection{Overlap}
This metric primarily assesses the overlap of graphic and text elements within the background. 
It consists of three major components. 
First, it measures the color change between the target area before and after rendering texts, considering whether the text is rendered against a background of the different color. 
Second, it calculates the percentage of overlap for each element, accounting for overlaps not only between text elements but also between text and graphics. Third, it assesses the percentage of the elements' area that extends beyond the canvas, measuring the extent to which text and other elements are appropriately contained within the background.

We transcribe the overlap principle to the prompt below.

\begin{lstlisting}
Overlapping elements are common in many designs and absent from others.
Less or proper overlapping might be considered aesthetically pleasing, but others are not.
Please consider the following points to evaluate the overlap.
1. The three types of overlap, the overlap of elements on text, the overlap of text on graphics, and the overlap of graphics on other graphics, are considered.
2. Hard-to-read text because of insufficient color contrast between a text and the background color is penalized.
3. The graphic design that includes elements extending past the boundaries is also penalized.
\end{lstlisting}

\subsection{White space}
This metric evaluates the appropriate amount of white space in a design and consists of five main components.
First, it measures the percentage of white space in the image, determined by the proportion except for graphic and text elements relative to the overall image area. 
Generally, a larger amount of natural white space is considered better. 
Second, it includes a metric that penalizes designs where certain parts of the image have excessively large areas of white space. 
In contrast to the first component, this metric helps ensure a balanced distribution of white space by penalizing areas with unnatural white space.
Third, the metric evaluates white space based on the distance between each element. This component considers the spacing between elements as a measure of white space adequacy.
Fourth, it assesses the variance in the distance between text elements. Consistent spacing between texts is considered aesthetically pleasing and indicates uniform white space.
Finally, the metric evaluates the white space at the edges of the image (i.e., border margin). An image lacking adequate white space at the edges is considered aesthetically poor.

We transcribe the white space principle to the prompt below.

\begin{lstlisting}
White space in graphic designs is fundamental for readability and aesthetics.
Element distance is also closely related to the principle of proximity, as elements placed near each other may appear to be related. 
White space also influences the overall design style; many modern designs use significant white space. White space 'trapped' between elements can also be distracting.
Evaluate the white space considering the following points.
1.A large ratio of white space that is not covered by design elements (e.g., graphics and tests) is preferred.
2. However, the graphic design with a too large region of empty white space on the image is undesirable.
3. The greater the distance between each element is preferred.
4. Uniformed vertical spacing of each text element is preferred.
5. Wider border margins for each element are preferred.
\end{lstlisting}

\section{Detailed prompts for GPT evaluation}
We create the prompt based on design principles described in \cite{o2014learning}.
Note that our prompts include a part of the prompt used in~\cite{jia2023cole}.
In Section~\ref{sec:pairwise}, we also conduct a pairwise evaluation that directly compares two graphic designs and determines the superior one.
Therefore, we describe the prompt for the absolute and pairwise evaluation here.

\noindent\textbf{The prompt of absolute evaluation}
\begin{lstlisting}
You are an autonomous AI Assistant who aids designers by providing insightful, objective, and constructive critiques of graphic design projects. Your goals are: "Deliver comprehensive and unbiased evaluations of graphic designs based on the following design principles."
Grade seriously. The range of scores is from 1 to 10. A flawless design can earn 10 points, a mediocre design can only earn 7 points, a design with obvious shortcomings can only earn 4 points, and a very poor design can only earn 1-2 points.
[A design principle]
If the output is too long, it will be truncated. Only respond in JSON format, no other information. Example of output for a better graphic design:{"score": 6, explanation: "(Please concisely explain the reason of the score.)"}
Please score the following images. [image]
\end{lstlisting}
Here, \texttt{[A design principle]} indicates the prompt described in Section \ref{sec:principle}, \texttt{[image]} indicates 
placeholder of the input image. 

\noindent\textbf{The prompt of pairwise evaluation}\par
\begin{lstlisting}
You are an autonomous AI Assistant who aids designers by providing insightful, objective, and constructive critiques of graphic design projects. Your goals are: "Deliver comprehensive and unbiased evaluations of graphic designs based on the following design principles."
[A design principle]
If the output is too long, it will be truncated. Only respond in JSON format, no other information. Example of output for a better graphic design (a):{"better_design": "a", explanation: "(Please concisely explain the reason of choice.)"}
If both images are the same quality, answer {"better_design": "both", explanation: "(Please concisely explain the reason of choice.)"}'
 Which of the following graphic designs has better quality regarding the above-described points? (a)[image] (b)[image]
\end{lstlisting}

\section{Pairwise evaluation}\label{sec:pairwise}
\begin{figure}[t]
  \centering
  \includegraphics[width=\linewidth]{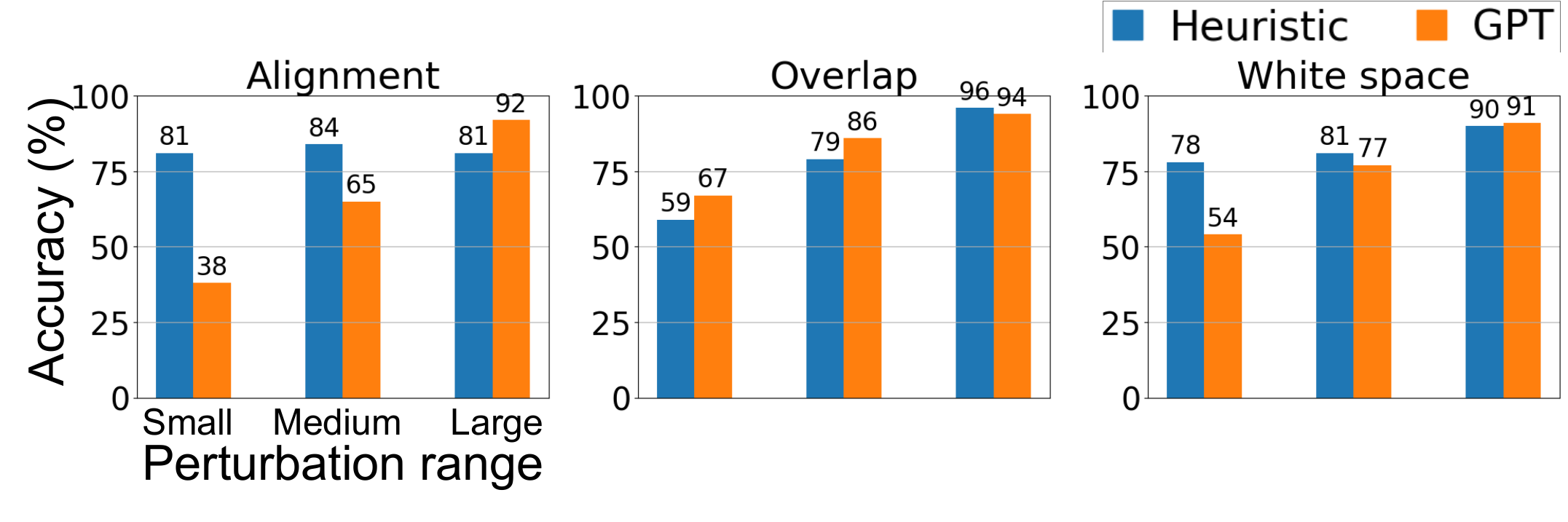}\\[-5mm]
  \caption{Comparison of pairwise evaluations.}
  \Description{This is a figure.}
  \label{fig:eval_comp}
\end{figure}

We also conduct the pairwise evaluation, which directly assesses which graphic design is better, inputting two images.
Two graphic designs are input to GPT-4o, and then GPT-4o answers which graphic design is better from the perspective of a specific design principle.
The choices are not limited to ``yes'' and ``no'' but also include ``not sure.'' Each pair is evaluated five times, and the final result is determined by voting.
We also conduct a similar evaluation task for humans to obtain the annotation.

We compare pairwise evaluation by GPT-4o with the heuristic evaluation.
In heuristic evaluation, we compare the scores of before and after perturbation and determine the better one.

Figure~\ref{fig:eval_comp} presents the pairwise evaluation results for each design principle. 
Across all metrics, the accuracy of GPT-4o increases progressively with the range of perturbation. 
For medium and large perturbations, GPT-4o performance is comparable to that of heuristic metrics, indicating that GPT-4o can distinguish between significantly poorer designs and others. 
However, GPT-4o performs worse than heuristic metrics for small perturbations in alignment and white space. 
There are two reasons for this. First, heuristic evaluation metrics are used to compare the two graphic designs before and after optimization. 
This is similar to the comparison of before and after perturbation designs. Second, as shown in the absolute evaluation, GPT-4o struggles to capture subtle differences in detailed design.

\begin{figure}[t]
  \centering
  \includegraphics[width=0.9\linewidth]{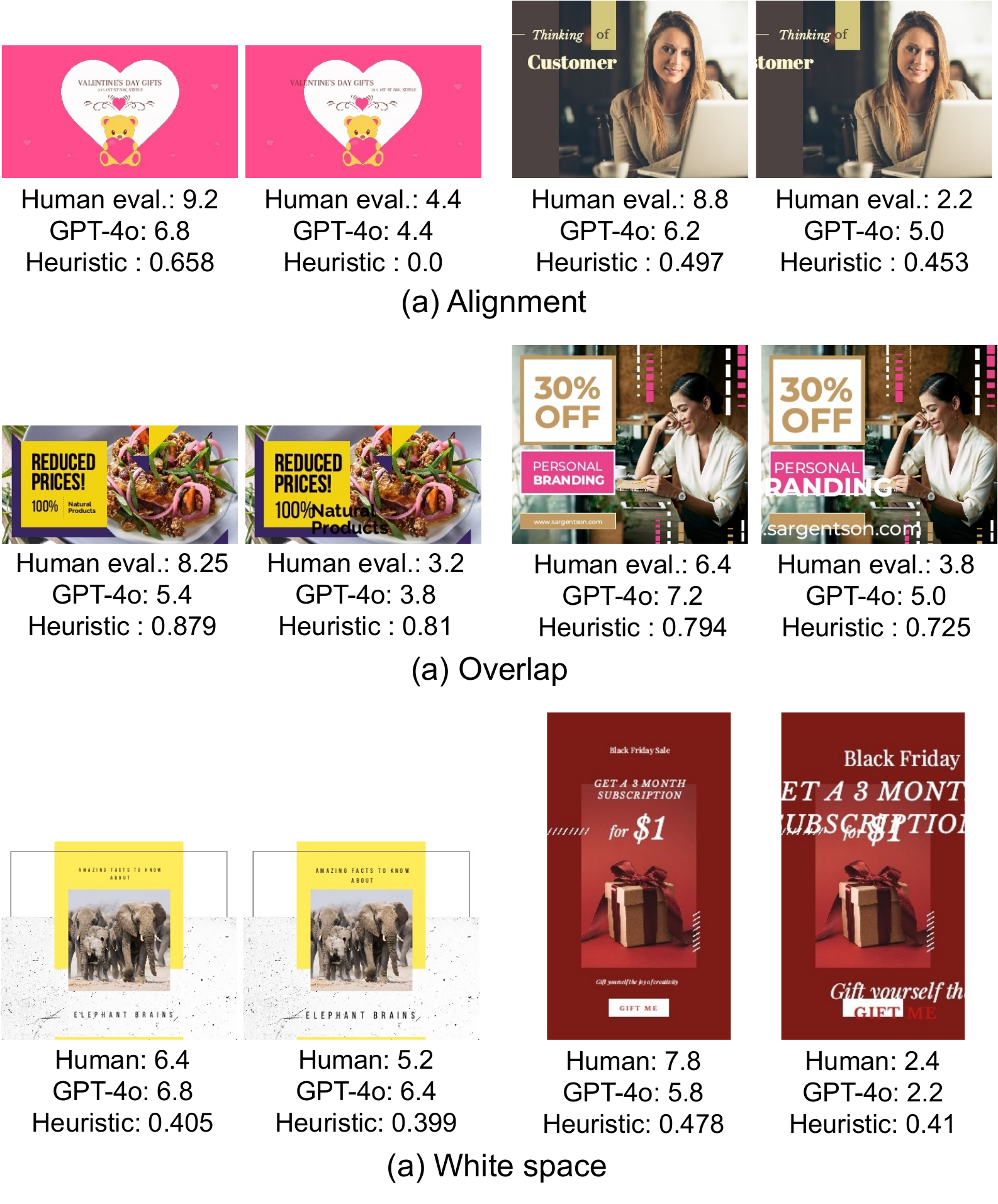} \\[-5mm]
  \caption{Examples of evaluation scores.}
  \Description{This is a figure.}
  \label{fig:score}
\end{figure}

\section{Examples of evaluation scores}
We show the examples of evaluation scores in Figure~\ref{fig:score}.
From these examples, GPT-4o scores are more similar to those of a heuristic evaluation. 
In contrast, heuristic evaluation can be used to compare the before and after perturbation; however, comparing the different graphic designs is difficult, as described in Section 5.1.

\end{document}